\documentclass[11pt]{article}
\usepackage{rotating}
\usepackage{acl2015}
\usepackage{times}
\usepackage{url}
\usepackage{latexsym}
\usepackage{tabularx}
\usepackage{makecell}
\usepackage{graphicx}
\usepackage{hyperref}
\usepackage{multirow}
\usepackage{multicol}
\usepackage{booktabs}
\usepackage{appendix}
\usepackage{CJKutf8}
\usepackage{amsmath}
\usepackage[section]{placeins}
\usepackage[hang,flushmargin]{footmisc} 
\usepackage[T1]{fontenc}
\usepackage{lipsum}
\usepackage{array}
\usepackage{stfloats} 
\usepackage{longtable}
\usepackage{xcolor}

\usepackage{fancyhdr}
\usepackage{caption}

\title{From LLM to Conversational Agent: A Memory Enhanced Architecture with Fine-Tuning of Large Language Models}

\author{Na Liu, Liangyu Chen, Xiaoyu Tian,  \\
\textbf{Wei Zou, Kaijiang Chen, Ming Cui}\\
Beike Inc., Beijing, China  \\
\texttt{\{liuna013, chenliangyu003, tianxiaoyu011,} \\
\texttt{zouwei026, chenkaijiang001, cuiming001\}@ke.com}}

\begin{document}
\maketitle

\begin{abstract}
This paper introduces RAISE (Reasoning and Acting through Scratchpad and Examples), an advanced architecture enhancing the integration of Large Language Models (LLMs) like GPT-4 into conversational agents. RAISE, an enhancement of the ReAct framework, incorporates a dual-component memory system, mirroring human short-term and long-term memory, to maintain context and continuity in conversations. It entails a comprehensive agent construction scenario, including phases like Conversation Selection, Scene Extraction, CoT Completion, and Scene Augmentation, leading to the LLMs Training phase. This approach appears to enhance agent controllability and adaptability in complex, multi-turn dialogues. Our preliminary evaluations in a real estate sales context suggest that RAISE has some advantages over traditional agents, indicating its potential for broader applications. This work contributes to the AI field by providing a robust framework for developing more context-aware and versatile conversational agents.
\end{abstract}

\section{Introduction}

The landscape of Artificial Intelligence (AI) is continuously evolving, with Large Language Models (LLMs) emerging as pivotal components in the advancement towards Artificial General Intelligence (AGI)\cite{ouyang2022training,wei2022emergentabilities,bubeck2023sparksofAGI}. These models, exemplified by GPT-4\cite{gpt4} and similar architectures, have demonstrated remarkable proficiency in a range of tasks, from conversation and reasoning to complex problem-solving in various domains. The versatility of LLMs has been further enriched by innovative prompting strategies and the integration of external tools, enhancing their capabilities beyond basic language processing.

However, a significant challenge in the realm of LLMs lies in their integration into conversational agents\cite{weng2023llmpowered,wang2023survey,sumers2023cognitive,xi2023rise} . While these models exhibit high levels of performance in isolated tasks, creating an agent that can sustain coherent, context-aware, and purpose-driven conversations remains an intricate endeavor. The need for a more sophisticated framework that leverages the strengths of LLMs while addressing their limitations in conversational settings has become increasingly apparent.

In response to this need, we introduce the RAISE (Reasoning and Acting through Scratchpad and Examples) architecture. RAISE represents a refined enhancement of the existing ReAct\cite{yao2023react} framework, specifically designed to augment the capabilities of conversational agents. This paper presents a detailed exploration of RAISE, highlighting its unique components and the benefits it offers in the development of conversational agents.

The cornerstone of RAISE is its incorporation of a dual-component memory system, analogous to the human brain's short-term and long-term memory functions. The Scratchpad component functions as a transient storage, capturing and processing key information and conclusions from recent interactions, akin to short-term memory. In parallel, the retrieval module operates as the agent's long-term memory, sourcing and incorporating examples relevant to the current conversational context. This enhanced memory mechanism can flexibly bolster the capabilities of conversational AI, and also provides a convenient interface for humans to customize and control the behavior of conversational AI systems.
\begin{figure*}[t]
\centering
\setlength{\abovecaptionskip}{0.1cm} 
\includegraphics[width=1\textwidth,trim={1cm 6cm 1cm 2cm} ]{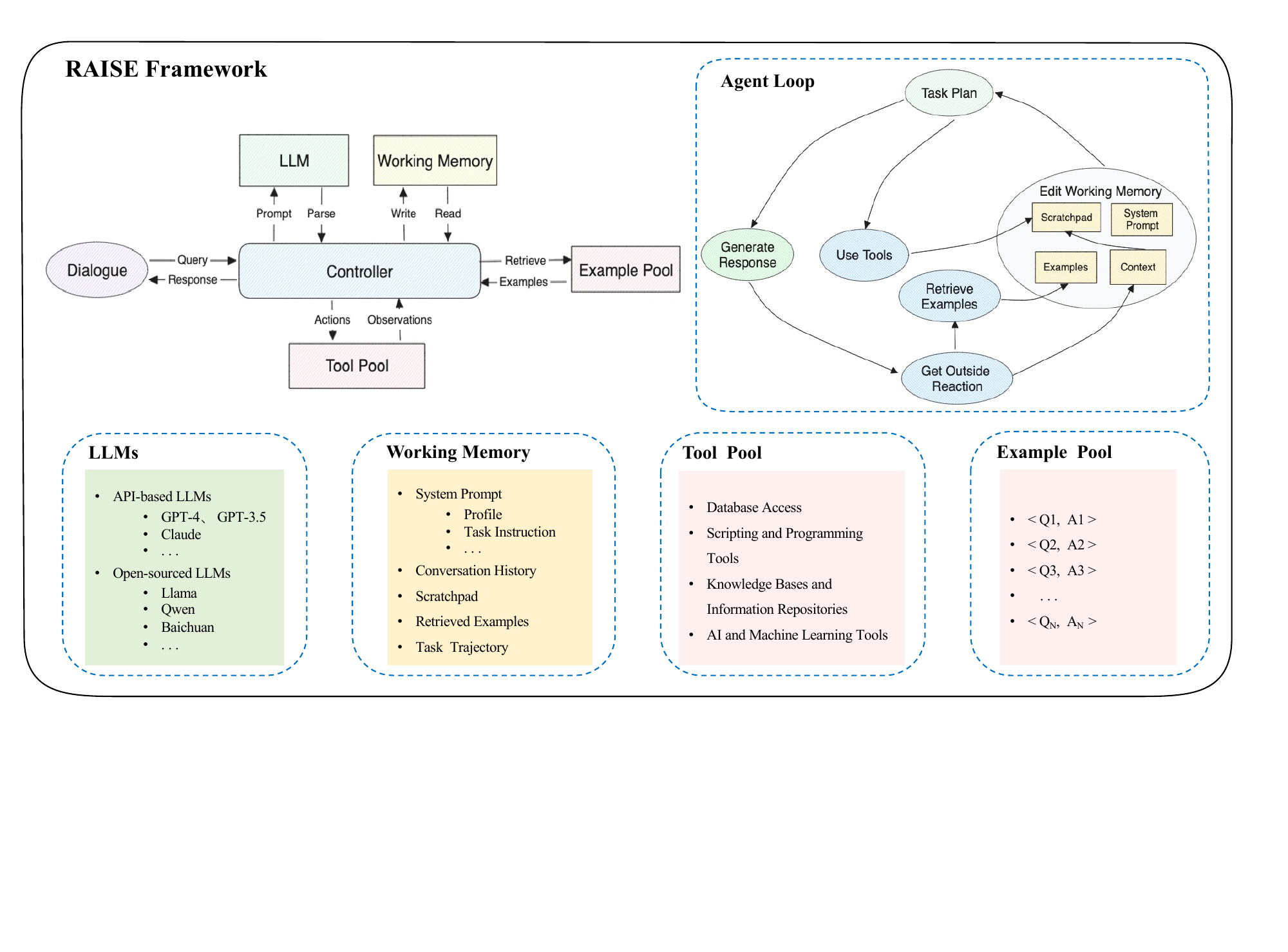}
\caption{The overview of RAISE.}
\label{raise_framework}
\end{figure*}

Furthermore, the RAISE architecture is founded on a comprehensive agent construction scenario, emphasizing the creation of conversational agents from scratch to ensure authenticity and relevance in real-world interactions. This paper delineates the RAISE methodology, encompassing a sequence of meticulously orchestrated phases. These include Conversation Selection, Scene Extraction, CoT (Chain of Thought)\cite{wei2022chain} Completion, and Scene Augmentation, all leading up to the pivotal LLMs Training phase. This structured approach is instrumental in developing agents that excel not only in language processing but also in contextual awareness and adaptability, catering to a spectrum of conversational dynamics.

Our experimental evaluations, conducted on a specialized in-house dataset focused on real estate sales, demonstrate the superiority of RAISE over conventional conversational agents. The results showcase RAISE's ability to handle complex, multi-turn conversations with enhanced context awareness and adaptability. While our experiments are centered on the real estate domain, the principles and methodologies underpinning RAISE are universally applicable, making it a versatile framework for various applications.

In summary, this paper presents the following contributions:
\begin{itemize}
\item 
We introduce RAISE, a refined enhancement of the ReAct framework, which utilizes scratchpad and retrieved examples to augment the agent's capabilities.
\item
We propose a fine-tuning scenario for Large Language Models (LLMs) within RAISE, which, compared to the use of prompts alone, not only enhances the controllability of the agent but also improves its effectiveness and efficiency.
\item
Through experiments conducted on our in-house dataset, we demonstrate RAISE's superiority as a conversational agent. While our experiments are concentrated on real estate sales, the underlying principles and methodologies of RAISE have wide-ranging applications and can be adapted to various domains, highlighting its versatility.
\end{itemize}

\section{Agent Framework}
\label{sec:agent_framework}
Inspired by ReAct \cite{yao2022react}, we introduce RAISE architecture, as shown in Figure \ref{raise_framework}. The architecture primarily encompasses the following components.

\subsection{Dialogue}
The dialogue module serves as the core interface for user-agent communication. It handles incoming user queries and delivers tailored responses formulated by the agent.

\subsection{LLMs}
As the agent's brain, the LLMs requires capabilities for perception, task-specific planning, tool usage, and summarization. These skills can be developed on the LLMs using either prompt engineering \cite{Crispino2023AgentIL} or fine-tuning methods \cite{zeng2023agenttuning}. Our study has conducted comparative experiments to stimulate these capabilities, utilizing models such as GPT-4 \cite{gpt4}, GPT-3.5 \cite{gpt3.5}, and Qwen-14B-Chat \cite{qwen}. This paper explores the strengths and limitations of each model in handling specific task types and provides concrete metrics for evaluating their performance.

\subsection{Memory}
The memory module in RAISE framework stores information perceived from its environment and facilitates the agent's future actions. The memory includes the following components:

\textbf{System Prompt}  \quad Includes profiles (detailing role identity, objectives, and behaviors), task instructions, tool descriptions, and few-shot learning elements for optimizing model performance. Flexibly designed, system prompt can either remain static or dynamically adjust to accommodate various stages of a dialogue and differing query types.

\textbf{Context}  \quad Includes conversation history and task trajectory. Conversation history records all query-response pairs within the dialogue, providing a complete context for more accurate agent perception. Task trajectory documents the decision-making trajectory, including plan designation, tool selection, and execution, guiding the agent's future planning.

\textbf{Scratchpad}  \quad Logs background information, knowledge generated by reasoning and observations from previous tool usage, essential for efficiency in multi-turn interactions.

\textbf{Examples}  \quad Comprises query-response pairs used for recalling relevant examples to supplement the model's and tools' knowledge gaps and to customize agent behavior and expression.

These four components collectively form the working memory of RAISE, with conversation history and scratchpad being dialogue-level, while examples and task trajectory are turn-level.

\subsection{Tool}
The tool module enriches LLMs after pretraining and Supervised Fine-Tuning (SFT) by integrating external knowledge sources and resources. This module incorporates a diverse array of tools, including but not limited to databases for data retrieval, APIs for system interactions, sophisticated recommendation systems, and collaborative frameworks involving other LLMs or agents. The description file for a tool typically needs to include the tool's name, its function, essential parameters, optional parameters, and may also include some usage examples. This descriptive file aids agents in better planning, tool selection, parameter generation for tools, and execution of those tools.

\subsection{Controller: Control Agent Loop}
The controller module connects the aforementioned modules through preset trigger conditions. Upon receiving a new query, the agent executes the loop of perception, planning, tool selection, and tool execution. The specific process is as follows.

\textbf{Memory Update}  \quad
At the beginning of a conversation, the \textit{Scratchpad} records the context of the dialogue, including user and agent roles, date, time, etc. 

During the conversation, each time a user query is received, the system will: (1) Add the user's query to the \textit{Conversation History}; (2) Recall top-\(k\) relevant examples from the \textit{Example Pool} for the current task, based on the historical and current query, using vector retrieval; (3) Update the current entity information in the \textit{Scratchpad} if the user's query contains a product link; (4) Update the agent's trajectory in the \textit{Task Memory} and the results of tool usage in the \textit{Scratchpad} during task execution; (5) Post-task completion, include the agent’s final output in the \textit{Conversation History}.

\begin{figure}[t]
\centering
\setlength{\abovecaptionskip}{0.1cm} 
\includegraphics[width= 0.95 \textwidth,trim={2cm 0cm 10cm 0cm}, clip]{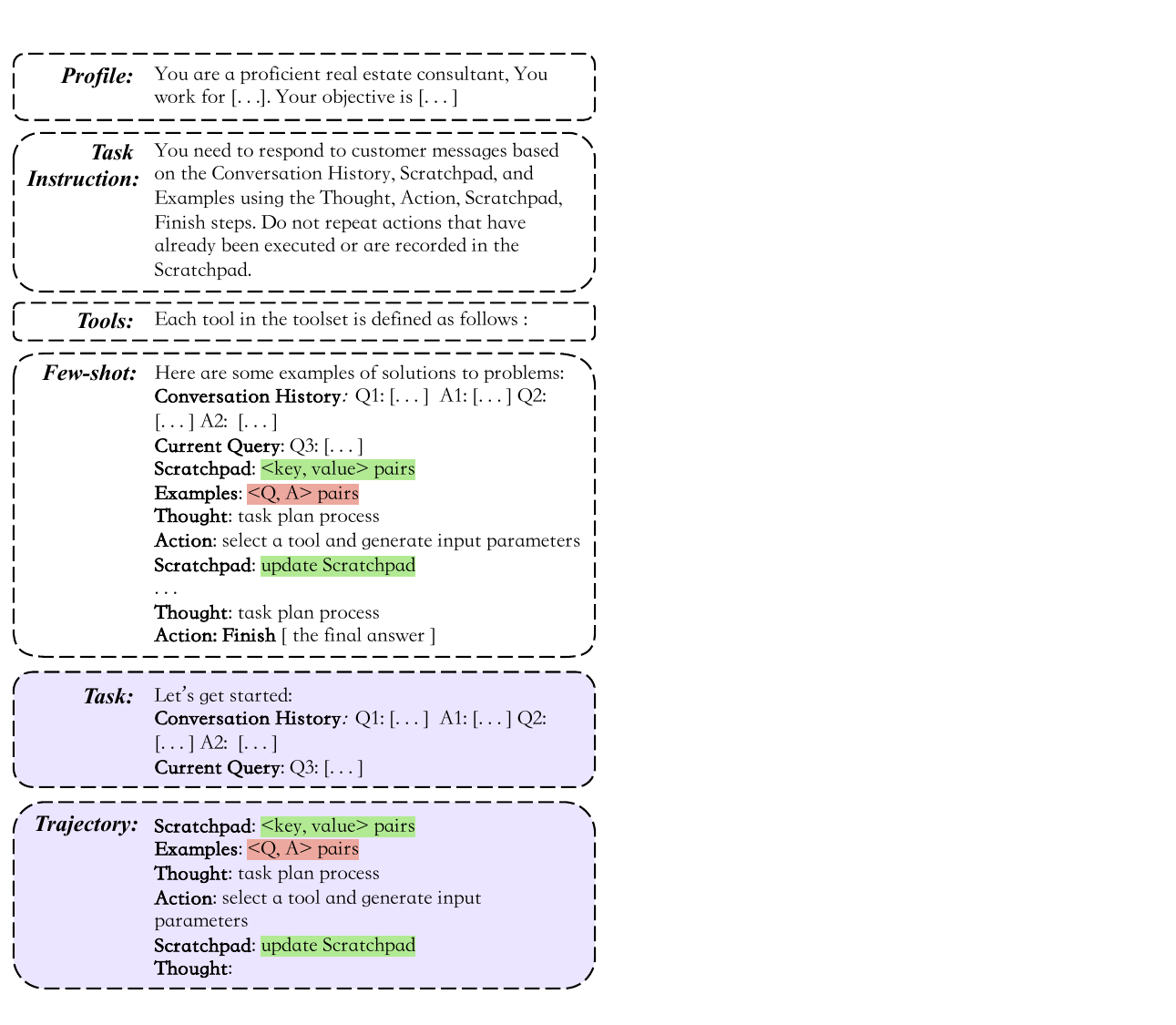}
\caption{Task Inference Prompt Template.}
\label{prompt_tempate}
\end{figure}

\textbf{Task Planning}  \quad
After collecting the above information, it is combined into a complete task inference prompt according to the designed template, as illustrated in Figure \ref{prompt_tempate}. An example of the complete prompt is available in Table \ref{raise prompt} of the Appendix. The \textit{LLM} utilizes the information within the prompt for perception and planning, subsequently outputting actions in accordance with the format outlined in the prompt. If an action involves invoking a tool, it should specify the tool's name and input parameters.

\textbf{Tool Execution}  \quad
This phase involves executing the tool selected in the previous step. The command for tool execution may either be directly output by the agent or correspond to a manually crafted function specific to each tool. The output of the execution is formatted as predetermined.

\textbf{Summary}  \quad
The agent, synthesizing all the information gathered from the environment, decides whether it can respond to the user's query. Termination criteria might include having gathered sufficient information, exceeding a preset number of loops, or encountering a system error. Upon meeting any of these conditions, the agent can proceed to summarize its findings and provide a response.

\section{Agent Tuning}

Section \ref{sec:agent_framework} presented the RAISE architecture, establishing a hardware base for agents in complex dialogues. This section shifts focus to software enhancements for RAISE, particularly activating LLMs as the agent's core. Despite the success of open-source LLMs in various tasks, studies \cite{liu2023agentbench} reveal their limitations in real-world scenarios, especially compared to GPT-3.5 \cite{gpt3.5} and GPT-4 \cite{gpt4}. Addressing this gap, this paper introduces a versatile finetuning method suitable for complex agent applications.

\subsection{Build Datasets}
\label{build_datasets}

\begin{figure*}[t]
\centering
\setlength{\abovecaptionskip}{0.1cm} 
\includegraphics[width=0.9\textwidth,trim={2cm 3cm 1cm 4cm} ]{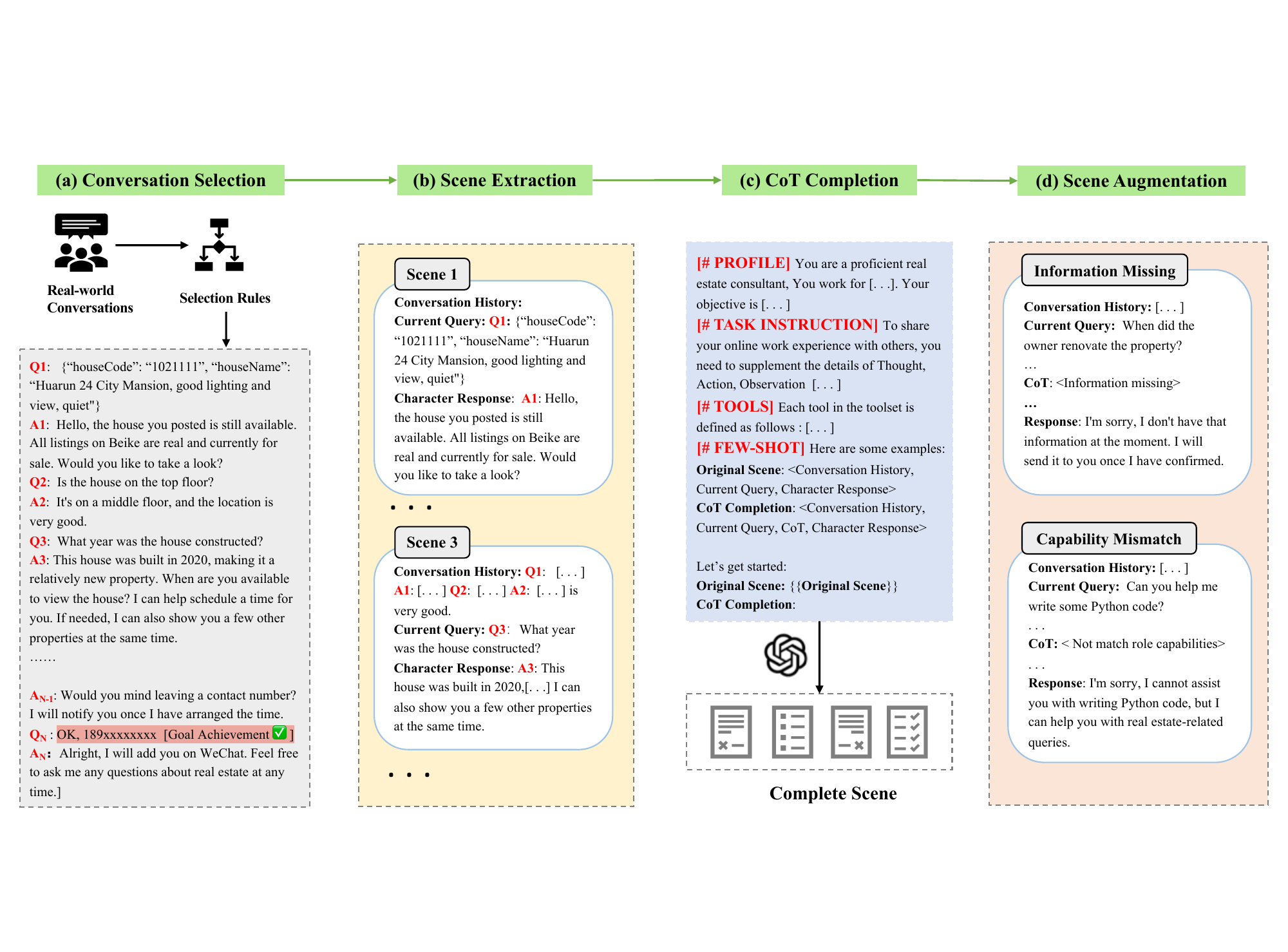}
\caption{Dataset Construction Pipeline for Agent Training.}
\label{build_datasets}
\end{figure*}

The creation of training data entails significant costs. Our objective is to finetune the model efficiently using a compact yet high-quality dataset that precisely aligns with specific role-based behavioral logic. The dataset must fulfill these criteria:

\textbf{Authenticity}  \quad It should closely mimic real-life scenarios.

\textbf{Diversity}  \quad The data should encompass a wide range of scenarios.

\textbf{High Quality}  \quad The data must have an accurate Chain of Thought (CoT) process, encompassing aspects like planning, tool utilization, and response formulation.

As shown in Figure \ref{build_datasets}, our proposed pipeline comprises several stages, including Conversation Selection, Scene Extraction, CoT Completion, and Scene Augmentation. The details of each stage are as follows:

\subsubsection{Conversation Selection}

To emulate specific roles in real scenarios, we start by filtering conversations from authentic dialogues based on criteria such as scene completion, a minimum number of dialogue turns, high conversation quality, and a threshold for user message ratio. These selected dialogues are then anonymized for further processing, as shown in Figure \ref{build_datasets}(a).

\subsubsection{Scene Extraction}

Each round of interaction serves as a segmentation point, dividing the previously selected dialogues into multiple samples, as shown in Figure \ref{build_datasets}(b). Each sample is an original scene (defined as $Scene_{origin}$). Assuming the user query at time $ i $ is defined as $ Q_i $ and the character's response as $A_i$, $Scene_{origin}$ comprises the following components: 

\small
\begin{align}
Scene^{origin}_{t} = \left\{
\begin{aligned}
& History_t: Q_1, A_1, \dots, Q_{t-1}, A_{t-1} \\
& Query_t: Q_t \\ 
& Response_t: A_t
\end{aligned}
\right.
\end{align}
\normalsize

Subsequently, to ensure diversity, we perform sampling based on dialogue turn counts and the intents behind user queries, resulting in a dataset rich in varied scene types.

\subsubsection{CoT Completion}

In refining the training data for the RAISE framework, the next phase involves enhancing the original scenes with a CoT process, which bridges the gap between user queries and character responses. This CoT process encompasses perception, planning, tool selection, and execution. Studies\cite{Nori2023CanGF} have demonstrated GPT-4's efficacy in generating high-quality CoT prompts for intricate scenarios. In this study, we initially utilize GPT-4 for automated generation, followed by meticulous manual validation of the output. To assist GPT-4 in consistently generating CoT processes, we incorporate additional elements such as predefined profiles, tools, and few-shot examples into the original scene, which collectively shape the construction of the prompt, as shown in Figure \ref{build_datasets}(c). The refined complete scene thus includes the following elements: 

\small
\begin{align}
Scene^{complete}_{t} = \left\{
\begin{aligned}
& History_t: Q_1, A_1, \dots, Q_{t-1}, A_{t-1} \\
& Query_t: Q_t \\ 
& CoT_t: Thought, Action, Observaton \\
& Response_t: A_t
\end{aligned}
\right.
\end{align}
\normalsize

\subsubsection{Scene Augmentation}

While the Scene Extraction phase ensured diversity through actual data sampling and the CoT Completion phase added the necessary CoT intricacies, two critical challenges still need addressing:

\textbf{Role Hallucination}  \quad LLMs, endowed with vast domain knowledge from pre-training and fine-tuning, exhibit extensive capabilities. However, if left unchecked, our trained agents might retain these broad skills, which could conflict with their intended functional roles. For example, an agent designed to provide sales services might erroneously possess skills like coding in Python or offering recipe advice. To counter this, we introduce specific scenarios that teach the agent its capability limits, essentially making it 'unlearn' the general abilities of LLMs within these defined contexts.

\begin{figure*}[t]
\centering
\setlength{\abovecaptionskip}{0.1cm} 
\includegraphics[width=0.95\textwidth,trim={2cm 4cm 2cm 4cm} ]{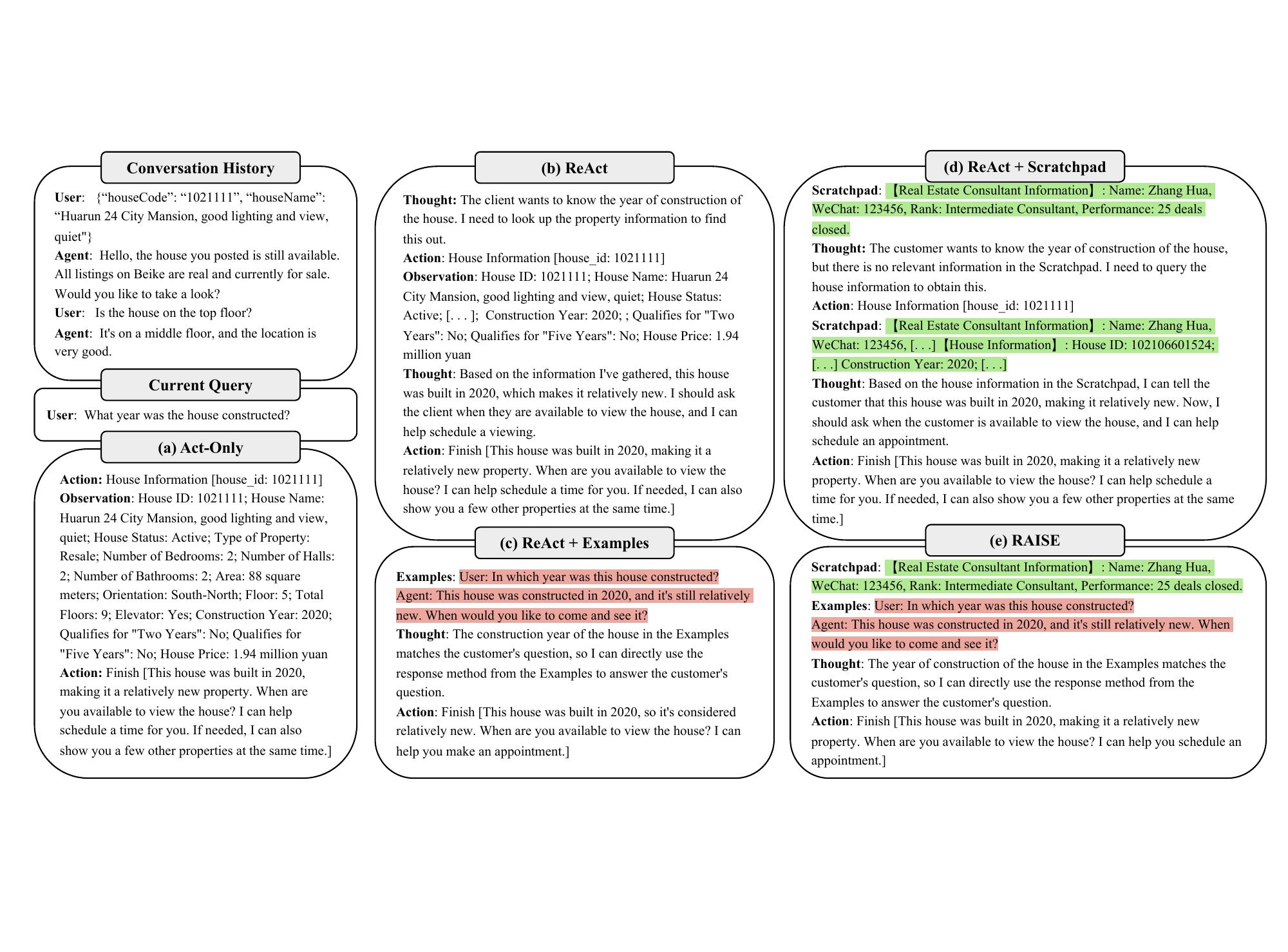}
\caption{Comparison of 5 Agent Frameworks.}
\label{five_methods}
\end{figure*}

\textbf{Knowledge Hallucination}  \quad
This phenomenon involves creating unrealistic or incorrect statements due to inadequate or misapplied knowledge acquired during pre-training. To mitigate this, we incorporate scenarios where the agent, despite tool utilization, still lacks essential factual knowledge, resulting in inability to respond accurately. In instances requiring factual accuracy, the agent should base its responses on knowledge acquired from its working memory or through tool interaction, rather than relying on its pre-trained database. 

To overcome these issues, we perform data augmentation on these two categories of data, which are not included in the real online datasets, as demonstrated in Figure \ref{build_datasets}(d).

\subsection{LLMs Training}

Following the previous phase, we have acquired a dataset characterized by both high quality and diversity. Each sample in this dataset, identified as $Scene_{complete}$ and combined with a $System \ Prompt$, constitutes a complete training $Sample$: 

\small
\begin{align}
Sample = \left\{
\begin{aligned}
& System Prompt: Profile, Instruction, \dots \\
& History_t: Q_1, A_1, \dots, Q_{t-1}, A_{t-1} \\
& Query_t: Q_t \\ 
& CoT_t: Thought, Action, Observaton \\
& Response_t: A_t
\end{aligned}
\right.
\end{align}
\normalsize

These instances are then processed into a format conducive for full-parameter fine-tuning of open-source LLMs. Our experiments have led to an encouraging discovery: by constructing a modest amount (<\(1K\)) of high-quality, representative data, the RAISE framework achieved impressive results.

\section{Experiments}

To demonstrate the effectiveness of the RAISE architecture and the fine-tuning method proposed in this paper in complex real-world scenarios, we conducted experiments in a real estate online Instant Messaging (IM) dialogue setting. In this scenario, the user is a customer inquiring about real estate purchases, and the agent assumes the role of a real estate consultant. A detailed introduction to the dataset, toolset, and the method of activating agent capabilities in LLMs is provided below.

\subsection{Datasets}

To ascertain the effectiveness of the RAISE framework, we conducted comparative evaluations with various architectures, including \textit{Act-Only}, \textit{ReAct}, \textit{ReAct+Scratchpad}, \textit{ReAct+Examples}, and \textit{RAISE}. This comparison aimed to ensure fair evaluation across different models, maintaining uniform dialogue scenarios and consistent additional knowledge in identical training samples across various datasets. The full trajectories for a single scenario under each architecture are depicted in Figure \ref{five_methods}.

Initially, we generated the \textit{ReAct} architecture dataset following the procedure outlined in Section \ref{build_datasets}. We then modified this data to create training sets for the other architectures. The methodologies applied were as follows:

\textbf{Act-Only}  \quad This architecture was formed by removing the 'thought' process from \textit{ReAct}, allowing for straightforward generation through coding.

\textbf{ReAct+Scratchpad}  \quad Building upon \textit{ReAct}, the initial \textit{Scratchpad} distribution comprised 20\% empty, 30\% partially informative, and 50\% fully informative content. The \textit{ReAct} data, when combined with varying output requirements, served as prompts for the regeneration of the CoT process using GPT-4.

\textbf{ReAct+Examples}  \quad Similar to \textit{ReAct+Scratchpad}, with the distribution of Examples set at 20\% empty, 30\% partially informative, and 50\% fully informative. The \textit{ReAct} data, merged with diverse output requirements, were reformulated into prompts for GPT-4-driven CoT regeneration.

\textbf{RAISE}  \quad  This model integrated aspects of both \textit{ReAct+Scratchpad} and \textit{ReAct+Examples}. The CoT process was similarly regenerated using GPT-4.

This structured approach enabled a thorough evaluation of each architectural element within the \textit{RAISE} framework, clearly demonstrating the incremental benefits introduced by each component. Following the outlined procedure, a total of 948 scenes were generated. Out of these, 100 were randomly selected to serve as the evaluation set, while the remaining 848 instances were used for fine-tuning the model.

\subsection{Tools}
Based on real estate online IM conversations, we have abstractly defined the following 12 tools, each including the tool name, input parameters, and functions:

\begin{itemize}
    \item \textbf{Real Estate Consultant Information} $[agent\_ucid]$: Retrieves the consultant's name, contact details, WeChat ID, ranking, performance metrics, and more.
\end{itemize}

\begin{itemize}
    \item \textbf{House Information $[house\_id]$}: Offers essential details about a property, including its size, price, floor level, school district presence, and renovation status.
\end{itemize}

\begin{itemize}
    \item \textbf{Community Information} $[resblock\_id]$: Provides insights into the community, covering aspects like green spaces, property management, building specifications, proximity to subway stations, schools, and medical facilities.
\end{itemize}

\begin{itemize}
    \item \textbf{House Layout Analysis} $[frame\_id]$: Analyzes the strengths and weaknesses of a property's layout.
\end{itemize}

\begin{itemize}
    \item \textbf{House Price Changes} $[house\_id]$: Tracks price fluctuations for a specific property.
\end{itemize}

\begin{itemize}
    \item \textbf{Community Price Changes} $[resblock\_id]$: Reports on average price trends within a particular community.
\end{itemize}
\begin{itemize}
    \item \textbf{Community Transactions} $[resblock\_id]$: Accesses recent transaction data from the same community.
\end{itemize}
\begin{itemize}
    \item \textbf{Tax Policy} $[city\_id]$: Updates on the latest tax regulations and implications.
\end{itemize}
\begin{itemize}
    \item \textbf{Loan Policy} $[city\_id]$: Delivers current information on loan policies.
\end{itemize}
\begin{itemize}
    \item \textbf{Market Analysis} $[city\_id]$: Provides up-to-date real estate market insights.
\end{itemize}
\begin{itemize}
    \item \textbf{Recommend Listings} $[Conversation \ History]$: Suggests property listings to customers based on their conversation history and inferred needs, including rationale for each recommendation.
\end{itemize}
\begin{itemize}
    \item \textbf{Value Report} $[house\_id]$: Generates a comprehensive value report card for a property, aimed at engaging customers and encouraging them to share their contact details.
\end{itemize}

\subsection{LLMs}

The models used in this study are as follows:

\textbf{OpenAI GPT} We utilized GPT-4 for generating all fine-tuning data and employed both GPT-3.5 and GPT-4 for prompting purposes. Both models were operated in ChatCompletion mode as of November 2023, with the temperature set to 0.5.

\textbf{Qwen-14B-Chat} An open-source conversational model from Alibaba Cloud, featuring 14 billion parameters, which has demonstrated exceptional performance in tool utilization \cite{Chen2023TEvalET}. Qwen-14B-Chat was used for both fine-tuning and prompting. The parameter configuration for the Supervised Fine-Tuning (SFT) is detailed in Table \ref{hyper-parameters}. The hyperparameter settings for the prompting and fine-tuning methods during inference are identical, also shown in Table \ref{hyper-parameters}. In the inference phase, we utilized an NVIDIA A100 GPU equipped with 80GB of memory, offering robust computational power and substantial memory capacity for efficient processing.

Another distinction between prompting and fine-tuning methods is the use of one-shot guidance in prompting for structured output generation, whereas fine-tuning omits this step. Complete prompts for various architectures are detailed in the appendix \ref{appendix}.

\begin{table}[t!]
\caption{Hyper-parameter settings for SFT and Inference}

\begin{center}
\begin{tabular}{l|c} 
\hline 
\multicolumn{2}{c}{\textbf{SFT Hyper-parameters}} \\
\hline
\textbf{Hyper parameter} & \textbf{Value} \\
\hline   
precision  & bfloat16 \\
model\_max\_length  & 4096 \\
epochs  & 3 \\
batch size  & 64 \\
learning rate  & 5e-6 \\
warmup ratio  & 0.03 \\
LR scheduler type  & cosine \\
\hline
\end{tabular}
\end{center}

\begin{center}
\begin{tabular}{l|c} 
\hline 
\multicolumn{2}{c}{\textbf{Inference Hyper-parameters}} \\
\hline
\textbf{Hyper parameter} & \textbf{Value} \\
\hline   
max\_new\_tokens  & 300 \\
top\_p  & 0.85 \\
temperature  & 0.5 \\
repetition\_penalty  & 1.1 \\
\hline
\end{tabular}
\end{center}

\label{hyper-parameters}
\end{table}

\renewcommand{\arraystretch}{1.3} 

\begin{table*}[htbp]
\caption{The evaluation criteria details of the defined metrics.}
\small
\begin{center}
\begin{tabular}{lccl}
\hline 
\textbf{Dimension} & \textbf{Metric} & \textbf{Score} & \textbf{Description} \\
\hline
\multirow{12}{*}{Quality} 
& \multirow{3}{*}{Specificity}
& 0 & Vague, general answer without specific information or details. \\
& & 1 & Provides some specifics, but lacks detail or full relevance to the question. \\
& & 2 & Directly addressing the user’s query with detailed and specific information. \\
\cline{2-4}
& \multirow{3}{*}{Factuality} 
& 0 & Contains false information, clearly contradicts facts. \\
& & 1 & Mostly accurate, with minor inaccuracies or oversights.\\
& & 2 & Completely accurate, all information is fact-checked. \\
\cline{2-4}
& \multirow{3}{*}{Coherence} 
& 0 & Logically disorganized, unrelated to prior content or overall topic. \\
& & 1 & Generally coherent, with some logical inconsistencies.\\
& & 2 & Very coherent, logically sound, closely aligned with the conversation topic. \\
\cline{2-4}
& \multirow{3}{*}{Naturalness} 
& 0 & Mechanical and unnatural, deviating from human conversational norms.\\
& & 1 & Imitates natural dialogue to an extent, but still somewhat stiff or unnatural.\\
& & 2 & Smooth and natural, akin to human dialogue, easily understood and accepted.\\
\hline
\multirow{3}{*}{Efficiency} 
& Plan Steps & - & Number of planning steps. \\ 
\cline{2-4}
& Action Steps & - & Number of action steps. \\
\cline{2-4}
& Inference Speed & - & The average time taken to process each user query, measured in seconds. \\
\hline
\end{tabular}
\end{center}
\label{evaluation}
\end{table*}

\subsection{Evaluation}
In the challenging landscape of human-computer dialogue systems, the evaluation of agent performance necessitates a nuanced approach \cite{liu2023agentbench,Wang2023ASO}. This is particularly pertinent when agents are tasked with engaging in direct conversations with human users, where the ultimate goal is to nurture a trust-based relationship. To achieve this, agents must exhibit a spectrum of qualities: they must be not only helpful and trustworthy but also responsive in a timely manner, and capable of understanding and articulating responses in a variety of contexts, akin to human interaction.
This paper delineates seven sophisticated metrics designed to rigorously assess both the quality and efficiency of agent responses. The specific metrics and their corresponding scoring criteria are detailed in Table \ref{evaluation}.

For assessing quality, we leverage human-centric annotation methods that closely replicate human evaluative standards. Meanwhile, the efficiency metrics are derived through a systematic statistical analysis, providing concrete, quantifiable insights.

\subsection{Ablation Study}

To demonstrate the effectiveness of the RAISE framework and fine-tuning method proposed in this paper, we conduct several ablation experiments in this section. The experiments are divided into two main aspects: (1) Comparative Analysis of Different Frameworks: We evaluate the performance of various frameworks under the same capability activation method, comparing their results using both the prompting and fine-tuning methods. The evaluation results are presented in Table \ref{mainresult_framework}. (2) Comparative Analysis of Different Capability Activation Methods: We compare the performance of the prompting method and fine-tuning method within the same framework, with the evaluation results presented in Table \ref{mainresult_method}.

It's important to note that the inference speed of the OpenAI GPT API is subject to platform and network variations, so this metric was omitted from our analysis. The inference environment for Qwen-14B-Chat was kept consistent, utilizing an A100 GPU with 80GB memory.

\renewcommand{\arraystretch}{1.1} 
\begin{table*}[t!]
\caption{The evaluation results of different frameworks}
\small
\begin{center}
\begin{tabular}{lccccc|ccc} 
\hline 
\textbf{Framework} & \textbf{Spec.} & \textbf{Fact.} & \textbf{Coher.} & \textbf{Nat.} & 
\textbf{Ov. Qual. Score} & 
\textbf{Plan Steps} & 
\textbf{Act. Steps} & 
\textbf{Inf. Speed(s)}  \\
\hline
\multicolumn{6}{l|}{\textbf{\textit{Prompting (GPT-4)}}} \\
Act-Only	&1.89 &1.66 &1.95 &1.87 &7.37 &- &1.29 &-\\
ReAct	&\textbf{1.98} &1.87	&1.93 &1.79	&7.57	&2	&1 &-\\
ReAct+Scratchpad &\textbf{1.98}	&1.88 &\textbf{1.99}	&1.65 &7.5	&1.97 &0.96 &-\\
ReAct+Examples &1.96	&1.87 &1.96	&\textbf{1.93} &\textbf{7.72}	&2.1 &1.1 &-\\
RAISE	&1.95 &\textbf{1.92}	&1.97 &1.85	&7.69 &\textbf{1.79} &\textbf{0.8} &-\\
\hline
\multicolumn{6}{l|}{\textbf{\textit{Fine-tuning (Qwen-14B-Chat)}}} \\
Act-Only &1.66 &1.71 &1.82 &1.92 &7.11	&-	&0.66 &\textbf{1.935} \\
ReAct	&1.88 &1.79	&1.93 &1.92	&7.52 &1.88 &0.88	&4.315\\
ReAct+Scratchpad &1.91 &1.81 &1.93 &1.96 &7.61	&1.6 &0.61 &3.833 \\
ReAct+Examples &\textbf{1.93} &1.82 &\textbf{1.96} &1.95 &7.66 &1.33	&0.33 &3.327 \\
RAISE	&1.87 &\textbf{1.9} &\textbf{1.96} &\textbf{1.98} &\textbf{7.71} &\textbf{1.26} &\textbf{0.26} &3.227 \\
\hline
\end{tabular}
\end{center}
\label{mainresult_framework}
\end{table*}

\begin{table*}[t!]
\caption{The evaluation results for prompting vs. fine-tuning}
\small
\begin{center}
\begin{tabular}{lccccc|ccc} 
\hline 
\textbf{Method} & \textbf{Spec.} & \textbf{Fact.} & \textbf{Coher.} & \textbf{Nat.} & 
\textbf{Ov. Qual. Score} & 
\textbf{Plan Steps} & 
\textbf{Act. Steps} \\
\hline
\textbf{\textit{RAISE}} & & & & & & & \\
Prompting (GPT-3.5)	&1.65 &1.72	&1.66 &1.67 &6.7 &2.13 &5 \\
Prompting (Qwen-14B-Chat) &1.69	&1.66 &1.68 &1.65	&6.68 &2.06	&1.2 \\
Prompting (GPT-4) &\textbf{1.95}	&\textbf{1.92} &\textbf{1.97} &1.85 &7.69	&1.79 &0.8 \\
Fine-tuning (Qwen-14B-Chat) &1.87 &1.9 &1.96 &\textbf{1.98	}&\textbf{7.71} &\textbf{1.26} &\textbf{0.26} \\
\hline
\textbf{\textit{ReAct+Scratchpad}} & & & & & & & & \\
Prompting (GPT-3.5)	&1.62 &1.57	&1.74 &1.55 &6.48	&2.19 &1.18 \\
Prompting (Qwen-14B-Chat) &1.68	&1.56	&1.71 &1.7 &6.65 &2.07 &1.09 \\
Prompting (GPT-4) &\textbf{1.98}	&\textbf{1.88} &\textbf{1.99} &1.65 &7.5 &1.97 &0.96 \\
Fine-tuning (Qwen-14B-Chat) &1.91 &1.81 &1.93 &\textbf{1.96	}&\textbf{7.61} &\textbf{1.6} &\textbf{0.61} \\
\hline
\end{tabular}
\end{center}
\label{mainresult_method}
\end{table*}

Upon analyzing the experimental outcomes, the following key conclusions emerge with respect to the efficacy and efficiency of different agent frameworks and methods:

\textbf{Framework Performance Ranking within the Same LLM} \quad The RAISE framework demonstrates superior performance, followed by ReAct+Examples, ReAct+Scratchpad, ReAct, and lastly, the Act-Only approach. This ranking indicates a clear gradient in effectiveness and efficiency, highlighting the incremental benefits of integrating additional elements like examples and scratchpads into the base ReAct model.

\textbf{Comparative Analysis of Capability Activation Methods within Identical Frameworks} \quad The fine-tuning approach outperforms the prompting method. This suggests that tailored training and customization of models to specific tasks or datasets result in more efficient and effective performance compared to using generalized prompt-based interactions.

In the following parts, we delve into detailed analyses of these findings, examining the implications and potential applications of each framework and methodology.

\textbf{Chain of Thought (CoT): A Catalyst for Enhanced Comprehension and Response Accurac} \quad CoT significantly boosts AI's ability to deeply comprehend and precisely respond to complex queries. Our experimental findings reaffirm the importance of CoT in complex tasks. For instance, in comparative experiments across different frameworks, agents employing the Act-Only method showed substantially lower performance compared to those incorporating CoT. These findings underscore the critical role of CoT in promoting AI models to deliver depth-oriented and logically coherent responses, particularly in scenarios requiring complex reasoning.

\textbf{RAISE Architecture: Dual Benefits of Efficiency from Scratchpad and Examples} \quad The RAISE architecture, by harmonizing Scratchpad and Example mechanisms, attains a dual advantage in processing efficiency and output quality. The application of Scratchpad significantly enhances the efficiency in handling complex tasks, while the utilization of Examples simultaneously bolsters the response's naturalness, specificity, and efficiency. This dual advantage positions the RAISE architecture as a suitable choice for scenarios demanding rapid, accurate, and naturally interactive responses.

\textbf{Fine-tuning: A Lever for Enhancing Agent Performance} \quad Utilizing diverse, high-quality datasets for fine-tuning helps to better align AI models with human behavioral logic, potentially leading to notable improvements in specific application areas. This approach excels in specialized tasks, offering a high degree of professionalism and customization. For instance, in the RAISE framework under fine-tuning, the overall quality score reached 7.71, and inference efficiency was optimized. These results validate the effectiveness of fine-tuning in delivering precise, human-like and efficient outcomes, particularly suitable for scenarios requiring customized solutions, such as online real estate services.

\textbf{Fine-tuning: Enhancing Cost-Efficiency and Speed during Agent Inference} \quad Although fine-tuning may require higher initial investments, its long-term benefits in operational efficiency and precision can offset these costs. In application, fine-tuned models often necessitate fewer computational resources, thereby reducing operational costs and accelerating response times. This cost-effectiveness, coupled with improved performance, makes fine-tuning a prudent choice for specific, resource-intensive tasks.

\textbf{Strategic Deployment of Language Agents: When to Choose Fine-tuning Over Prompting} \quad The decision to opt for fine-tuning or prompting hinges on the specific requirements of the application. Fine-tuning offers superior performance and efficiency in specialized domains but may involve higher initial costs and training needs. In contrast, prompting is more flexible in handling a wide range of queries but may slightly lag behind fine-tuned models in specificity and stability. Strategic decision-making in this context involves balancing these factors against the specific needs of the application, budget constraints, and performance expectations.

\section{Related Work}

The exploration and advancements in AI agents have captivated the AI research community for some time. Defined as artificial entities capable of perceiving their surroundings, making decisions, and executing actions \cite{zalta1995stanford,barandiaran2009defining}, AI agents represent a significant stride in artificial intelligence.

The advent of Large Language Models (LLMs) has been a pivotal development, often regarded as a step towards the realization of Artificial General Intelligence (AGI) \cite{ouyang2022training,wei2022emergentabilities,bubeck2023sparksofAGI}. In recent years, there has been an influx of studies proposing intricate LLM-based architectures for AI agents \cite{weng2023llmpowered,wang2023survey,sumers2023cognitive,xi2023rise}. These architectures are crucial in enabling agents to navigate complex dialogue scenarios and effectively apply their acquired knowledge.

This body of work primarily revolves around two core aspects:

\begin{itemize}
\item \textbf{Planning}: Central to the functionality of dialogue agents is the concept of Chain-of-Thought (CoT) reasoning. This involves eliciting logical rationales via CoT prompts, as explored in \cite{wei2022chain,wang2023selfconsistency,zhou2023least}. However, integrating this reasoning effectively into dialogues remains challenging. The ReAct framework \cite{yao2023react} presents an approach that guides LLMs in reasoning before planning actions, addressing this issue.

\item \textbf{Tool Use}: Another critical facet is the ability of LLMs to utilize external tools and resources. Studies such as \cite{schick2023toolformer,li2023apibank,shen2023hugginggpt} have demonstrated the proficiency of LLMs in leveraging external tools and APIs. Moreover, the capacity to extract and integrate knowledge from external sources has been further exemplified by projects like WebGPT \cite{nakano2022webgpt} and ExpeL \cite{zhao2023expel}.
\end{itemize}

In addition to these areas, several works have focused on broader algorithmic frameworks for LLM-based agents.\cite{xie2023openagents,pan2023kwaiagents,sumers2023cognitive,ruan2023tptu,kong2023tptu,li2023modelscope} On the other hand, specific dialogue agents have also been a focal point.\cite{shao2023character,wang2023rolellm,chen2023chatcot,chae2023dialogue,hong2023zero}. 
The fine-tuning of LLMs within agents is another critical area, with works like \cite{zeng2023agenttuning,chen2023fireact} exploring this aspect.

These developments underscore the growing complexity and capabilities of LLM-based AI agents, highlighting both the challenges and the innovations shaping the field.

\section{Conclusions and Future work}

This study introduces RAISE, an advanced architecture enhancing Long Language Models (LLMs) like GPT-4 for conversational agents. Building on the ReAct framework, RAISE integrates a dual-component memory system, improving dialogue context retention and continuity. We also propose a fine-tuning method within RAISE, which enhances agent controllability and efficiency, particularly in real estate sales, though applicable in various domains.

However, the study has limitations, including potential hallucination issues and challenges in handling complex logic problems, necessitating further research. Despite these limitations, RAISE presents a promising advancement in adaptable, context-aware conversational agents, offering a foundation for future developments in artificial intelligence.

\bibliographystyle{acl}
\bibliography{acl2015}

\begin{thebibliography}{}

\bibitem[\protect\citename{Bai \bgroup et al.\egroup }2023]{qwen}
Jinze Bai, Shuai Bai, Yunfei Chu, Zeyu Cui, Kai Dang, Xiaodong Deng, Yang Fan, Wenbin Ge, Yu~Han, Fei Huang, Binyuan Hui, Luo Ji, Mei Li, Junyang Lin, Runji Lin, Dayiheng Liu, Gao Liu, Chengqiang Lu, Keming Lu, Jianxin Ma, Rui Men, Xingzhang Ren, Xuancheng Ren, Chuanqi Tan, Sinan Tan, Jianhong Tu, Peng Wang, Shijie Wang, Wei Wang, Shengguang Wu, Benfeng Xu, Jin Xu, An~Yang, Hao Yang, Jian Yang, Shusheng Yang, Yang Yao, Bowen Yu, Hongyi Yuan, Zheng Yuan, Jianwei Zhang, Xingxuan Zhang, Yichang Zhang, Zhenru Zhang, Chang Zhou, Jingren Zhou, Xiaohuan Zhou, and Tianhang Zhu.
\newblock 2023.
\newblock Qwen technical report.
\newblock {\em arXiv preprint arXiv:2309.16609}.

\bibitem[\protect\citename{Barandiaran \bgroup et al.\egroup }2009]{barandiaran2009defining}
Xabier~E Barandiaran, Ezequiel Di~Paolo, and Marieke Rohde.
\newblock 2009.
\newblock Defining agency: Individuality, normativity, asymmetry, and spatio-temporality in action.
\newblock {\em Adaptive Behavior}, 17(5):367--386.

\bibitem[\protect\citename{Bubeck \bgroup et al.\egroup }2023]{bubeck2023sparksofAGI}
S.~Bubeck, V.~Chandrasekaran, R.~Eldan, et~al.
\newblock 2023.
\newblock Sparks of artificial general intelligence: Early experiments with gpt-4.
\newblock {\em CoRR}.
\newblock arXiv:2303.12712.

\bibitem[\protect\citename{Chae \bgroup et al.\egroup }2023]{chae2023dialogue}
Hyungjoo Chae, Yongho Song, Kai Tzu-iunn Ong, Taeyoon Kwon, Minjin Kim, Youngjae Yu, Dongha Lee, Dongyeop Kang, and Jinyoung Yeo.
\newblock 2023.
\newblock Dialogue chain-of-thought distillation for commonsense-aware conversational agents.
\newblock {\em arXiv preprint arXiv:2310.09343}.

\bibitem[\protect\citename{Chen \bgroup et al.\egroup }2023a]{chen2023fireact}
Baian Chen, Chang Shu, Ehsan Shareghi, Nigel Collier, Karthik Narasimhan, and Shunyu Yao.
\newblock 2023a.
\newblock Fireact: Toward language agent fine-tuning.
\newblock {\em arXiv preprint arXiv:2310.05915}.

\bibitem[\protect\citename{Chen \bgroup et al.\egroup }2023b]{Chen2023TEvalET}
Zehui Chen, Weihua Du, Wenwei Zhang, Kuikun Liu, Jiangning Liu, Miao Zheng, Jingming Zhuo, Songyang Zhang, Dahua Lin, Kai Chen, and Feng Zhao.
\newblock 2023b.
\newblock T-eval: Evaluating the tool utilization capability step by step.

\bibitem[\protect\citename{Chen \bgroup et al.\egroup }2023c]{chen2023chatcot}
Zhipeng Chen, Kun Zhou, Beichen Zhang, Zheng Gong, Wayne~Xin Zhao, and Ji-Rong Wen.
\newblock 2023c.
\newblock Chatcot: Tool-augmented chain-of-thought reasoning on$\backslash$$\backslash$chat-based large language models.
\newblock {\em arXiv preprint arXiv:2305.14323}.

\bibitem[\protect\citename{Crispino \bgroup et al.\egroup }2023]{Crispino2023AgentIL}
Nicholas Crispino, Kyle Montgomery, Fankun Zeng, Dawn Song, and Chenguang Wang.
\newblock 2023.
\newblock Agent instructs large language models to be general zero-shot reasoners.
\newblock {\em ArXiv}, abs/2310.03710.

\bibitem[\protect\citename{Hong \bgroup et al.\egroup }2023]{hong2023zero}
Joey Hong, Sergey Levine, and Anca Dragan.
\newblock 2023.
\newblock Zero-shot goal-directed dialogue via rl on imagined conversations.
\newblock {\em arXiv preprint arXiv:2311.05584}.

\bibitem[\protect\citename{Kong \bgroup et al.\egroup }2023]{kong2023tptu}
Yilun Kong, Jingqing Ruan, Yihong Chen, Bin Zhang, Tianpeng Bao, Shiwei Shi, Guoqing Du, Xiaoru Hu, Hangyu Mao, Ziyue Li, et~al.
\newblock 2023.
\newblock Tptu-v2: Boosting task planning and tool usage of large language model-based agents in real-world systems.
\newblock {\em arXiv preprint arXiv:2311.11315}.

\bibitem[\protect\citename{Li \bgroup et al.\egroup }2023a]{li2023modelscope}
Chenliang Li, Hehong Chen, Ming Yan, Weizhou Shen, Haiyang Xu, Zhikai Wu, Zhicheng Zhang, Wenmeng Zhou, Yingda Chen, Chen Cheng, et~al.
\newblock 2023a.
\newblock Modelscope-agent: Building your customizable agent system with open-source large language models.
\newblock {\em arXiv preprint arXiv:2309.00986}.

\bibitem[\protect\citename{Li \bgroup et al.\egroup }2023b]{li2023apibank}
Minghao Li, Feifan Song, Bowen Yu, Haiyang Yu, Zhoujun Li, Fei Huang, and Yongbin Li.
\newblock 2023b.
\newblock Apibank: A benchmark for tool-augmented llms.
\newblock {\em arXiv preprint}.

\bibitem[\protect\citename{Liu \bgroup et al.\egroup }2023]{liu2023agentbench}
Xiao Liu, Hao Yu, Hanchen Zhang, Yifan Xu, Xuanyu Lei, Hanyu Lai, Yu~Gu, Hangliang Ding, Kaiwen Men, Kejuan Yang, et~al.
\newblock 2023.
\newblock Agentbench: Evaluating llms as agents.
\newblock {\em arXiv preprint arXiv:2308.03688}.

\bibitem[\protect\citename{Nakano \bgroup et al.\egroup }2022]{nakano2022webgpt}
Reiichiro Nakano, Jacob Hilton, Suchir Balaji, Jeff Wu, Long Ouyang, Christina Kim, Christopher Hesse, Shantanu Jain, Vineet Kosaraju, William Saunders, Xu~Jiang, Karl Cobbe, Tyna Eloundou, Gretchen Krueger, Kevin Button, Matthew Knight, Benjamin Chess, and John Schulman.
\newblock 2022.
\newblock Webgpt: Browser-assisted question-answering with human feedback.
\newblock {\em arXiv preprint}.

\bibitem[\protect\citename{Nori \bgroup et al.\egroup }2023]{Nori2023CanGF}
Harsha Nori, Yin~Tat Lee, Sheng Zhang, Dean Carignan, Richard Edgar, Nicolo Fusi, Nicholas King, Jonathan Larson, Yuanzhi Li, Weishung Liu, Renqian Luo, Scott~Mayer McKinney, Robert~Osazuwa Ness, Hoifung Poon, Tao Qin, Naoto Usuyama, Chris White, and Eric Horvitz.
\newblock 2023.
\newblock Can generalist foundation models outcompete special-purpose tuning? case study in medicine.
\newblock {\em ArXiv}, abs/2311.16452.

\bibitem[\protect\citename{OpenAI}2023a]{gpt3.5}
OpenAI.
\newblock 2023a.
\newblock Chatgpt: Optimizing language models for dialogue.
\newblock Blog post.

\bibitem[\protect\citename{OpenAI}2023b]{gpt4}
OpenAI.
\newblock 2023b.
\newblock Gpt-4 technical report.
\newblock Blog post.

\bibitem[\protect\citename{Ouyang \bgroup et al.\egroup }2022]{ouyang2022training}
Long Ouyang, Jeff Wu, Xu~Jiang, Diogo Almeida, Carroll~L. Wainwright, Pamela Mishkin, Chong Zhang, Sandhini Agarwal, Katarina Slama, Alex Ray, John Schulman, Jacob Hilton, Fraser Kelton, Luke Miller, Maddie Simens, Amanda Askell, Peter Welinder, Paul Christiano, Jan Leike, and Ryan Lowe.
\newblock 2022.
\newblock Training language models to follow instructions with human feedback.

\bibitem[\protect\citename{Pan \bgroup et al.\egroup }2023]{pan2023kwaiagents}
Haojie Pan, Zepeng Zhai, Hao Yuan, Yaojia Lv, Ruiji Fu, Ming Liu, Zhongyuan Wang, and Bing Qin.
\newblock 2023.
\newblock Kwaiagents: Generalized information-seeking agent system with large language models.
\newblock {\em arXiv preprint arXiv:2312.04889}.

\bibitem[\protect\citename{Ruan \bgroup et al.\egroup }2023]{ruan2023tptu}
Jingqing Ruan, Yihong Chen, Bin Zhang, Zhiwei Xu, Tianpeng Bao, Guoqing Du, Shiwei Shi, Hangyu Mao, Xingyu Zeng, and Rui Zhao.
\newblock 2023.
\newblock Tptu: Task planning and tool usage of large language model-based ai agents.
\newblock {\em arXiv preprint arXiv:2308.03427}.

\bibitem[\protect\citename{Schick \bgroup et al.\egroup }2023]{schick2023toolformer}
Timo Schick, Jane Dwivedi-Yu, Roberto Dessì, Roberta Raileanu, Maria Lomeli, Luke Zettlemoyer, Nicola Cancedda, and Thomas Scialom.
\newblock 2023.
\newblock Toolformer: Language models can teach themselves to use tools.
\newblock {\em arXiv preprint}.

\bibitem[\protect\citename{Shao \bgroup et al.\egroup }2023]{shao2023character}
Yunfan Shao, Linyang Li, Junqi Dai, and Xipeng Qiu.
\newblock 2023.
\newblock Character-llm: A trainable agent for role-playing.
\newblock {\em arXiv preprint arXiv:2310.10158}.

\bibitem[\protect\citename{Shen \bgroup et al.\egroup }2023]{shen2023hugginggpt}
Yongliang Shen, Kaitao Song, Xu~Tan, Dongsheng Li, Weiming Lu, and Yueting Zhuang.
\newblock 2023.
\newblock Hugginggpt: Solving ai tasks with chatgpt and its friends in huggingface.
\newblock {\em arXiv preprint arXiv:2303.17580}.

\bibitem[\protect\citename{Sumers \bgroup et al.\egroup }2023]{sumers2023cognitive}
Theodore Sumers, Shunyu Yao, Karthik Narasimhan, and Thomas~L Griffiths.
\newblock 2023.
\newblock Cognitive architectures for language agents.
\newblock {\em arXiv preprint arXiv:2309.02427}.

\bibitem[\protect\citename{Wang \bgroup et al.\egroup }2023a]{wang2023survey}
Lei Wang, Chen Ma, Xueyang Feng, Zeyu Zhang, Hao Yang, Jingsen Zhang, Zhiyuan Chen, Jiakai Tang, Xu~Chen, Yankai Lin, et~al.
\newblock 2023a.
\newblock A survey on large language model based autonomous agents.
\newblock {\em arXiv preprint arXiv:2308.11432}.

\bibitem[\protect\citename{Wang \bgroup et al.\egroup }2023b]{Wang2023ASO}
Lei Wang, Chengbang Ma, Xueyang Feng, Zeyu Zhang, Hao ran Yang, Jingsen Zhang, Zhi-Yang Chen, Jiakai Tang, Xu~Chen, Yankai Lin, Wayne~Xin Zhao, Zhewei Wei, and Ji~rong Wen.
\newblock 2023b.
\newblock A survey on large language model based autonomous agents.
\newblock {\em ArXiv}, abs/2308.11432.

\bibitem[\protect\citename{Wang \bgroup et al.\egroup }2023c]{wang2023selfconsistency}
Xuezhi Wang, Jason Wei, Dale Schuurmans, Quoc~V Le, Ed~H. Chi, Sharan Narang, Aakanksha Chowdhery, and Denny Zhou.
\newblock 2023c.
\newblock Self-consistency improves chain of thought reasoning in language models.
\newblock In {\em Proceedings of ICLR}.

\bibitem[\protect\citename{Wang \bgroup et al.\egroup }2023d]{wang2023rolellm}
Zekun~Moore Wang, Zhongyuan Peng, Haoran Que, Jiaheng Liu, Wangchunshu Zhou, Yuhan Wu, Hongcheng Guo, Ruitong Gan, Zehao Ni, Man Zhang, et~al.
\newblock 2023d.
\newblock Rolellm: Benchmarking, eliciting, and enhancing role-playing abilities of large language models.
\newblock {\em arXiv preprint arXiv:2310.00746}.

\bibitem[\protect\citename{Wei \bgroup et al.\egroup }2022a]{wei2022emergentabilities}
J.~Wei, Y.~Tay, R.~Bommasani, et~al.
\newblock 2022a.
\newblock Emergent abilities of large language models.
\newblock {\em Trans. Mach. Learn. Res.}

\bibitem[\protect\citename{Wei \bgroup et al.\egroup }2022b]{wei2022chain}
Jason Wei, Xuezhi Wang, Dale Schuurmans, Maarten Bosma, Brian Ichter, Fei Xia, Ed~H. Chi, Quoc~V Le, and Denny Zhou.
\newblock 2022b.
\newblock Chain of thought prompting elicits reasoning in large language models.
\newblock In {\em Proceedings of NeurIPS}.

\bibitem[\protect\citename{Weng}2023]{weng2023llmpowered}
L.~Weng.
\newblock 2023.
\newblock Llm-powered autonomous agents.

\bibitem[\protect\citename{Xi \bgroup et al.\egroup }2023]{xi2023rise}
Zhiheng Xi, Wenxiang Chen, Xin Guo, Wei He, Yiwen Ding, Boyang Hong, Ming Zhang, Junzhe Wang, Senjie Jin, Enyu Zhou, et~al.
\newblock 2023.
\newblock The rise and potential of large language model based agents: A survey.
\newblock {\em arXiv preprint arXiv:2309.07864}.

\bibitem[\protect\citename{Xie \bgroup et al.\egroup }2023]{xie2023openagents}
Tianbao Xie, Fan Zhou, Zhoujun Cheng, Peng Shi, Luoxuan Weng, Yitao Liu, Toh~Jing Hua, Junning Zhao, Qian Liu, Che Liu, et~al.
\newblock 2023.
\newblock Openagents: An open platform for language agents in the wild.
\newblock {\em arXiv preprint arXiv:2310.10634}.

\bibitem[\protect\citename{Yao \bgroup et al.\egroup }2022]{yao2022react}
Shunyu Yao, Jeffrey Zhao, Dian Yu, Nan Du, Izhak Shafran, Karthik Narasimhan, and Yuan Cao.
\newblock 2022.
\newblock React: Synergizing reasoning and acting in language models.
\newblock {\em arXiv preprint arXiv:2210.03629}.

\bibitem[\protect\citename{Yao \bgroup et al.\egroup }2023]{yao2023react}
Shunyu Yao, Jeffrey Zhao, Dian Yu, Nan Du, Izhak Shafran, Karthik Narasimhan, and Yuan Cao.
\newblock 2023.
\newblock React: Synergizing reasoning and acting in language models.
\newblock {\em arXiv preprint}.

\bibitem[\protect\citename{Zalta \bgroup et al.\egroup }1995]{zalta1995stanford}
Edward~N Zalta, Uri Nodelman, Colin Allen, and John Perry.
\newblock 1995.
\newblock Stanford encyclopedia of philosophy.

\bibitem[\protect\citename{Zeng \bgroup et al.\egroup }2023]{zeng2023agenttuning}
Aohan Zeng, Mingdao Liu, Rui Lu, Bowen Wang, Xiao Liu, Yuxiao Dong, and Jie Tang.
\newblock 2023.
\newblock Agenttuning: Enabling generalized agent abilities for llms.
\newblock {\em arXiv preprint arXiv:2310.12823}.

\bibitem[\protect\citename{Zhao \bgroup et al.\egroup }2023]{zhao2023expel}
Andrew Zhao, Daniel Huang, Quentin Xu, Matthieu Lin, Yong-Jin Liu, and Gao Huang.
\newblock 2023.
\newblock Expel: Llm agents are experiential learners.
\newblock {\em arXiv preprint arXiv:2308.10144}.

\bibitem[\protect\citename{Zhou \bgroup et al.\egroup }2023]{zhou2023least}
Denny Zhou, Nathanael Schärli, Le~Hou, Jason Wei, Nathan Scales, Xuezhi Wang, Dale Schuurmans, Claire Cui, Olivier Bousquet, Quoc~V Le, and Ed~H. Chi.
\newblock 2023.
\newblock Least-to-most prompting enables complex reasoning in large language models.
\newblock In {\em Proceedings of ICLR}.

\end{thebibliography}

\appendix
\section{Appendix}
\label{appendix}

The complete prompts used by the five agent frameworks during inference are shown in Table \ref{raise prompt} to Table \ref{react_e prompt}.

\begin{table*}[!b]
\centering
\caption{The prompt used for RAISE}
\begin{tabular}{m{15cm}}
\hline
\textcolor{blue}{You are a} proficient real estate consultant working for Beike Zhaofang, a company that provides real estate brokerage services. The company's value lies in assisting buyers to find their ideal homes. It envisions becoming a quality residential platform serving 300 million families, and its mission is to be a dignified service provider, contributing to a better living experience. Your objective, during online chat interactions, is to answer clients' questions, attract them to purchase properties, and encourage them to add you on WeChat or meet in person. \\\\
\textcolor{blue}{You need to} respond to client queries using the steps of \textbf{Scratchpad}, \textbf{Examples}, \textbf{Thought}, \textbf{Action}, \textbf{Observation}, \textbf{Finish}, based on historical conversations and the client's questions. Avoid repeating actions that have been used before.\\\\
Each tool in the \textcolor{blue}{toolset} is defined as follows:\\
\{tool descriptions\}
\\\\
Here is an \textcolor{blue}{example}: \textit{(Omit in the fine-tuning method.)}\\
\textbf{Conversation History}: User: {“houseCode”: “1021111”, “houseName”: “Huarun 24 City Mansion, good lighting and view, quiet"} \\
\textbf{Current Query}: What year was the house constructed?\\
\textbf{Scratchpad}: [Real Estate Consultant Information]: Name: Zhang Hua, WeChat: 123456, Rank: Intermediate Consultant, Performance: 25 deals closed. \\
\textbf{Examples}: User: In which year was this house constructed?
Agent: This house was constructed in 2020, and it's still relatively new. When would you like to come and see it?\\
\textbf{Thought}: The year of construction of the house in the Examples matches the customer's question, so I can directly use the response method from the Examples to answer the customer's question.
\textbf{Action: Finish} [This house was built in 2020, making it a relatively new property. When are you available to view the house? I can help you schedule an appointment.]\\\\
\textcolor{blue}{Let’s get started}:\\
\textbf{Conversation History}: \{Conversation History\} \\
\textbf{Current Query}: \{Current Query\}\\
\hline
\end{tabular}
\label{raise prompt}
\end{table*}

\begin{table*}[h]
\centering
\caption{The prompt used for Act-Only}
\begin{tabular}{m{16cm}}
\hline
\textcolor{blue}{You are a }proficient real estate consultant working for Beike Zhaofang, a company that provides real estate brokerage services. The company's value lies in assisting buyers to find their ideal homes. It envisions becoming a quality residential platform serving 300 million families, and its mission is to be a dignified service provider, contributing to a better living experience. Your objective, during online chat interactions, is to answer clients' questions, attract them to purchase properties, and encourage them to add you on WeChat or meet in person. \\\\
\textcolor{blue}{You need to }respond to client queries using the steps of \textbf{Action}, \textbf{Observation}, \textbf{Finish}, based on historical conversations and the client's questions. Avoid repeating actions that have been used before.\\\\
Each tool in the \textcolor{blue}{toolset} is defined as follows:\\
(1) Real Estate Consultant Information $[agent\_ucid]$: Retrieves the consultant's name, contact details, WeChat ID, ranking, performance metrics, and more.\\
(2) House Information $[house\_id]$: Offers essential details about a property, including its size, price, floor level, school district presence, and renovation status.\\
(3) Community Information $[resblock\_id]$: Provides insights into the community, covering aspects like green spaces, property management, building specifications, proximity to subway stations, schools, and medical facilities.\\
(4) House Layout Analysis $[frame\_id]$: Analyzes the strengths and weaknesses of a property's layout.\\
(5) House Price Changes $[house\_id]$: Tracks price fluctuations for a specific property.\\
(6) Community Price Changes $[resblock\_id]$: Reports on average price trends within a particular community.\\
(7) unity Transactions $[resblock\_id]$: Accesses recent transaction data from the same community.\\
(8) Tax Policy $[city\_id]$: Updates on the latest tax regulations and implications.\\
(9) Loan Policy $[city\_id]$: Delivers current information on loan policies.\\
(10)Market Analysis $[city\_id]$: Provides up-to-date real estate market insights.\\
(11) Recommend Listings $[Conversation \ History]$: Suggests property listings to customers based on their conversation history and inferred needs, including rationale for each recommendation.\\
(12) Value Report $[house\_id]$: Generates a comprehensive value report card for a property, aimed at engaging customers and encouraging them to share their contact details.\\\\
Here is an \textcolor{blue}{example}: \textit{(Omit in the fine-tuning method.)}\\
\textbf{Conversation History}: User: {“houseCode”: “1021111”, “houseName”: “Huarun 24 City Mansion, good lighting and view, quiet"} \\
\textbf{Current Query}: What year was the house constructed?\\
\textbf{Action}: House Information [house\_id: 1021111]\\
\textbf{Observation}: House ID: 1021111; House Name: Huarun 24 City Mansion, good lighting and view, quiet; House Status: Active; Type of Property: Resale; Number of Bedrooms: 2; Number of Halls: 2; Number of Bathrooms: 2; Area: 88 square meters; Orientation: South-North; Floor: 5; Total Floors: 9; Elevator: Yes; Construction Year: 2020; Qualifies for "Two Years": No; Qualifies for "Five Years": No; House Price: 1.94 million yuan\\
\textbf{Action: Finish} \([\)This house was built in 2020, making it a relatively new property. When are you available to view the house? I can help schedule a time for you. If needed, I can also show you a few other properties at the same time.\(]\)\\\\
\textcolor{blue}{Let’s get started}:\\
\textbf{Conversation History}: \{Conversation History\} \\
\textbf{Current Query}: \{Current Query\}\\
\hline
\end{tabular}
\label{act-only prompt}
\end{table*}

\begin{table*}[h]
\centering
\caption{The prompt used for ReAct}
\begin{tabular}{m{16cm}}
\hline
\textcolor{blue}{You are a} proficient real estate consultant working for Beike Zhaofang, a company that provides real estate brokerage services. The company's value lies in assisting buyers to find their ideal homes. It envisions becoming a quality residential platform serving 300 million families, and its mission is to be a dignified service provider, contributing to a better living experience. Your objective, during online chat interactions, is to answer clients' questions, attract them to purchase properties, and encourage them to add you on WeChat or meet in person. \\\\
\textcolor{blue}{You need to} respond to client queries using the steps of \textbf{Thought}, \textbf{Action}, \textbf{Observation}, \textbf{Finish}, based on historical conversations and the client's questions. Avoid repeating actions that have been used before.\\\\
Each tool in the \textcolor{blue}{toolset} is defined as follows:\\
\{tool descriptions\}over
\\\\
Here is an \textcolor{blue}{example}: \textit{(Omit in the fine-tuning method.)}\\
\textbf{Conversation History}: User: {“houseCode”: “1021111”, “houseName”: “Huarun 24 City Mansion, good lighting and view, quiet"} \\
\textbf{Current Query}: What year was the house constructed?\\

\textbf{Thought}: The client wants to know the year of construction of the house. I need to look up the property information to find this out.
\textbf{Action}: House Information [house\_id: 1021111]\\
\textbf{Observation}: House ID: 1021111; House Name: Huarun 24 City Mansion, good lighting and view, quiet; House Status: Active; Type of Property: Resale; Number of Bedrooms: 2; Number of Halls: 2; Number of Bathrooms: 2; Area: 88 square meters; Orientation: South-North; Floor: 5; Total Floors: 9; Elevator: Yes; Construction Year: 2020; Qualifies for "Two Years": No; Qualifies for "Five Years": No; House Price: 1.94 million yuan
\textbf{Thought}: Based on the information I've gathered, this house was built in 2020, which makes it relatively new. I should ask the client when they are available to view the house, and I can help schedule a viewing.\\
\textbf{Action: Finish} \([\)This house was built in 2020, making it a relatively new property. When are you available to view the house? I can help schedule a time for you. If needed, I can also show you a few other properties at the same time.\(]\)\\\\
\textcolor{blue}{Let’s get started}:\\
\textbf{Conversation History}: \{Conversation History\} \\
\textbf{Current Query}: \{Current Query\}\\
\hline
\end{tabular}
\label{react prompt}
\end{table*}

\begin{table*}[h]
\centering
\caption{The prompt used for ReAct+Scratchpad}
\begin{tabular}{m{16cm}}
\hline
\textcolor{blue}{You are a} proficient real estate consultant working for Beike Zhaofang, a company that provides real estate brokerage services. The company's value lies in assisting buyers to find their ideal homes. It envisions becoming a quality residential platform serving 300 million families, and its mission is to be a dignified service provider, contributing to a better living experience. Your objective, during online chat interactions, is to answer clients' questions, attract them to purchase properties, and encourage them to add you on WeChat or meet in person. \\\\
\textcolor{blue}{You need to} respond to client queries using the steps of \textbf{Scratchpad}, \textbf{Thought}, \textbf{Action}, \textbf{Finish}, based on historical conversations and the client's questions. Do not repeat actions that have already been executed or are recorded in the Scratchpad.\\\\
Each tool in the \textcolor{blue}{toolset} is defined as follows:\\
\{tool descriptions\}
\\\\
Here is an \textcolor{blue}{example}: \textit{(Omit in the fine-tuning method.)}\\
\textbf{Conversation History}: User: {“houseCode”: “1021111”, “houseName”: “Huarun 24 City Mansion, good lighting and view, quiet"} \\
\textbf{Current Query}: What year was the house constructed?\\

\textbf{Scratchpad}:[Real Estate Consultant Information]: Name: Zhang Hua, WeChat: 123456, Rank: Intermediate Consultant, Performance: 25 deals closed. \\
\textbf{Thought}: The customer wants to know the year of construction of the house, but there is no relevant information in the Scratchpad. I need to query the house information to obtain this.\\
\textbf{Action}: House Information [house\_id: 1021111]\\
\textbf{Scratchpad}: [Real Estate Consultant Information]: Name: Zhang Hua, WeChat: 123456, Rank: Intermediate Consultant, Performance: 25 deals closed.  [House Information]: House ID: 1021111; House Name: Huarun 24 City Mansion, good lighting and view, quiet; House Status: Active; Type of Property: Resale; Number of Bedrooms: 2; Number of Halls: 2; Number of Bathrooms: 2; Area: 88 square meters; Orientation: South-North; Floor: 5; Total Floors: 9; Elevator: Yes; Construction Year: 2020; Qualifies for "Two Years": No; Qualifies for "Five Years": No; House Price: 1.94 million yuan.\\
\textbf{Thought}: Based on the house information in the Scratchpad, I can tell the customer that this house was built in 2020, making it relatively new. Now, I should ask when the customer is available to view the house, and I can help schedule an appointment. \\
\textbf{Action: Finish} \([\)This house was built in 2020, making it a relatively new property. When are you available to view the house? I can help schedule a time for you. If needed, I can also show you a few other properties at the same time.\(]\)\\\\
\textcolor{blue}{Let’s get started}:\\
\textbf{Conversation History}: \{Conversation History\} \\
\textbf{Current Query}: \{Current Query\}\\
\hline
\end{tabular}
\label{react_s prompt}
\end{table*}

\begin{table*}[h]
\centering
\caption{The prompt used for ReAct+Examples}
\begin{tabular}{m{16cm}}
\hline
\textcolor{blue}{You are a} proficient real estate consultant working for Beike Zhaofang, a company that provides real estate brokerage services. The company's value lies in assisting buyers to find their ideal homes. It envisions becoming a quality residential platform serving 300 million families, and its mission is to be a dignified service provider, contributing to a better living experience. Your objective, during online chat interactions, is to answer clients' questions, attract them to purchase properties, and encourage them to add you on WeChat or meet in person. \\\\
\textcolor{blue}{You need to} respond to client queries using the steps of \textbf{Examples}, \textbf{Thought}, \textbf{Action}, \textbf{Observation}, \textbf{Finish}, based on historical conversations and the client's questions. Avoid repeating actions that have been used before.\\\\
Each tool in the \textcolor{blue}{toolset} is defined as follows:\\
\{tool descriptions\}
\\\\
Here is an \textcolor{blue}{example}: \textit{(Omit in the fine-tuning method.)}\\
\textbf{Conversation History}: User: {“houseCode”: “1021111”, “houseName”: “Huarun 24 City Mansion, good lighting and view, quiet"} \\
\textbf{Current Query}: What year was the house constructed?\\
\textbf{Examples}: User: In which year was this house constructed? 
Agent: This house was constructed in 2020, and it's still relatively new. When would you like to come and see it?\\
\textbf{Thought}: The construction year of the house in the Examples matches the customer's question, so I can directly use the response method from the Example to answer the customer's question.\\
\textbf{Action: Finish} [This house was built in 2020, so it's considered relatively new. When are you available to view the house? I can help you make an appointment.]\\\\
\textcolor{blue}{Let’s get started}:\\
\textbf{Conversation History}: \{Conversation History\} \\
\textbf{Current Query}: \{Current Query\}\\
\hline
\end{tabular}
\label{react_e prompt}
\end{table*}

\end{document}